\documentclass{article}
\usepackage{arxiv}
\usepackage[utf8]{inputenc} 
\usepackage[T1]{fontenc}    
\usepackage{hyperref}       
\usepackage{url}            
\usepackage{booktabs}       
\usepackage{amsfonts}       
\usepackage{nicefrac}       
\usepackage{microtype}      
\usepackage{lipsum}

\usepackage{url,hyperref,lineno,microtype}
\usepackage[onehalfspacing]{setspace}

\makeatletter
\newcommand*{\addFileDependency}[1]{
  \typeout{(#1)}
  \@addtofilelist{#1}
  \IfFileExists{#1}{}{\typeout{No file #1.}}
}
\makeatother

\usepackage{amsmath,amssymb,amsthm}

\usepackage{amsfonts}       
\usepackage{tikz}
\usepackage{bm}
\usepackage{float}

\usepackage{verbatim}
\usepackage{hhline}

\usepackage{graphicx}
\usepackage{caption}

\graphicspath{{./figures/}}
\newcommand{\V}[1]{\textbf{#1}}

\DeclareMathOperator*{\argmax}{\arg\!\max}
\newtheorem{theorem}{Theorem}[section]

\allowdisplaybreaks

\def\firstAuthorLast{Sample {et~al.}} 
\def\Authors{Ariel Jaffe\,$^{1,*}$, Yuval Kluger\,$^{2}$, Ofir Lindenbaum\,$^{1}$, Jonathan Patsenker\,$^{1}$, Erez Peterfreund\,$^{3}$, Stefan Steinerberger\,$^{4}$}

\def\Address{$^{1}$Program in Applied Mathematics, Yale University, New Haven, USA\\
$^{2}$Department of Pathology, Yale School of Medicine, New Haven, USA\\
$^3$School of Computer Science and Engineering, The Hebrew University, Jerusalem, Israel\\
$^{4}$Department of Mathematics, University of Washington, Seattle, WA 98195, USA}

\def\corrAuthor{Corresponding Author}

\def\corrEmail{a.jaffe@yale.edu}

\begin{document}

\title{The spectral underpinning of word2vec} 

\author{\Authors} 


\maketitle

\begin{abstract}
Word2vec introduced by Mikolov \textit{et al.} (2013) is a word embedding method that is widely used in natural language processing. Despite its success and frequent use, a strong theoretical justification is still lacking. The main contribution of our paper is to propose a rigorous analysis of the highly nonlinear functional of word2vec. Our results suggest that word2vec may be primarily driven by an underlying spectral method. This insight may open the door to obtaining provable guarantees for word2vec. 
We support these findings by numerical simulations. 
One fascinating open question is whether the nonlinear properties of word2vec that are not captured by the spectral method are beneficial and, if so, by what mechanism.



\end{abstract}

\section{Introduction }

Word2vec was introduced by Mikolov \textit{et al.} \cite{mikolov2013efficient} as an unsupervised scheme for embedding words based on text corpora. We will try to introduce the idea in the simplest possible terms
and refer to \cite{mikolov2013efficient,goldberg2014word2vec, grover2016node2vec} for the way it is usually presented.
Let $\left\{x_1, x_2, \dots, x_n\right\}$ be a set of elements for which we aim to compute a numerical representation. These may be words, documents, or nodes in a graph. 
Our input consists of an $n \times n$ matrix $P$ with non-negative elements $P_{ij}$, which
encode, by a numerical value,  the relationship between the set $\{x_i\}_{i=1}^n$ and a set of context elements $\{c_j\}_{j=1}^n$. The meaning of contexts is determined by the specific application, where in most cases the set of contexts is equal to the set of elements  (i.e. $c_i= x_i$ for any $i\in \{1,\ldots,n\}$) \cite{goldberg2014word2vec}.

The larger the value of $P_{ij}$, the larger the connection between $x_i$ and $c_j$. For example, such a connection can be quantified by the probability that a word appears in the same sentence as another word. Based on $P$, Mikolov defined an energy function which depends on two sets of vector representations   $\{w_1,\ldots,w_n\}$ and $\{v_1,\ldots,v_n\}$. Maximizing the functional with respect to these sets yields  $\{w_1^*,\ldots,w_n^*\}$ and $\{v_1^*,\ldots,v_n^*\}$ which can serve as a low dimensional representations for the words and contexts respectively. Ideally, this embedding should encapsulate the relations captured by the matrix $P$.


Assuming a uniform prior over the $n$ elements, the energy function $L: \mathbb{R}^n \times \mathbb{R}^n \rightarrow \mathbb{R}$, introduced by Mikolov et. al \cite{mikolov2013efficient} can be written as
\begin{equation}\label{eq:skip_gram_function}
L(w, v) = \left\langle w, Pv\right\rangle - \sum_{i=1}^{n} \log\left( \sum_{j=1}^{n} \exp(w_i v_j)\right).
\end{equation} 
The exact relation between \eqref{eq:skip_gram_function} and the formulation in \cite{mikolov2013distributed} appears in supplementary material.
 Word2vec is based on maximizing this expression over all $(w,v) \in \mathbb{R}^n \times \mathbb{R}^n$
$$ (w^*, v^*) = \argmax_{(w,v)} L(w,v).$$
There is no reason to assume that the maximum is unique. It has been observed that if $x_i$ and $x_j$ are similar elements in the data set (namely, words that frequently appear in the same sentence), then $v_i^*, v_j^*$ or $w_i^*, w_j^*$ tend to have similar numerical values. Thus, the values $\left\{w_1^*, \dots, w_n^*\right\}$ are useful for 
embedding  $\left\{x_1, \dots, x_n\right\}$.
 One could also try to maximize the symmetric loss that arises from enforcing $w=v$ and is given by $L:\mathbb{R}^n \rightarrow \mathbb{R}$
\begin{equation}\label{eq:symmetric_loss} 
L(w) = \left\langle w, Pw\right\rangle - \sum_{i=1}^{n} \log\left( \sum_{j=1}^{n} \exp(w_i w_j)\right).
\end{equation}
In Section \ref{sec:Examples} we show that the symmetric functional yields a meaningful embedding for various datasets. Here, the interpretation of the functional is straight-forward: we wish to pick $w \in \mathbb{R}^n$ in a way that makes $\left\langle w, Pw \right\rangle$ large. Assuming $P$ is diagonalizable, 
this is achieved for $w$ that is a linear combination of the leading eigenvectors. 
At the same time, the exponential function places a penalty over large entries in $w$.  


Our paper initiates a
rigorous study of the energy functional $L(w)$, however, we emphasize that a complete description of the energy landscape $L(w)$ remains an interesting open problem.
We also emphasize that our analysis has direct implications for computational aspects as well: for instance, if one were interested in maximizing the nonlinear functional, the maximum of its linear approximation (which is easy to compute) is a natural starting point.
A simple example is shown in Figure 1: the underlying dataset contains $200$ points in $\mathbb{R}^{10}$ where
the first $100$ points are drawn from a Gaussian distribution, and the second $100$ points are drawn from a second Gaussian distribution. The matrix $P$ is the row-stochastic matrix induced by a
Gaussian kernel $K_{ij} = \exp(-\|x_i - x_j\|^2/\alpha)$ where $\alpha$ is a scaling parameter discussed in Section \ref{sec:Examples}. We observe that, up to scaling, the maximizer of the energy functional (black) is well approximated
by the spectral methods introduced below.

\section{Motivation and Related works } 

Optimizing over energy functions such as \eqref{eq:skip_gram_function} to obtain vector embeddings is done for various applications, such as words \cite{mikolov2013distributed}, documents \cite{le2014distributed} and graphs \cite{narayanan2017graph2vec}. 
Surprisingly, very few works addressed the analytic aspects of optimizing over the word2vec functional. 
Hashimoto et. al. \cite{hashimoto2016word} derived a relation between word2vec and stochastic neighbor embedding \cite{hinton2003stochastic}. 
Cotterell et. al. \cite{cotterell2017explaining} showed that when $P$ is sampled according to a multinomial distribution, optimizing over \eqref{eq:skip_gram_function} is equivalent to exponential family PCA \cite{collins2002generalization}. 
If the number of elements is large, optimizing over \eqref{eq:skip_gram_function} becomes impractical. As an efficient alternative, Mikolov et. al. \cite{mikolov2013distributed} suggested a variation 
based on negative sampling. 
Levy and Goldberg \cite{levy2014neural} showed that if the embedding dimension is sufficiently high, then optimizing over the negative sampling functional suggested in \cite{mikolov2013distributed} is equivalent to factorizing the shifted Pointwise Mutual Information matrix. 
This work was extended in \cite{qiu2018network}, where similar results were derived for additional embedding algorithms such as  \cite{grover2016node2vec,perozzi2014deepwalk,tang2015line}.
Decomposition of the PMI matrix was also justified by Arora et. al. \cite{Arora2015RandomWO}, based on a generative random walk model.
A different approach was introduced by
Landgraf \cite{landgraf2017word2vec},  
that related the negative sampling loss function to logistic PCA. 

In this work,  
we focus on approximating the highly nonlinear word2vec functional by Taylor expansion. We show that in the regime of embedding vectors with small magnitude, the functional can be approximated by the spectral decomposition of the matrix $P$.
This draws a natural connection between word2vec and classical, 
spectral embedding methods such as \cite{belkin2003laplacian,coifman2006diffusion}. By rephrasing
word2vec as a spectral method in the ‘small vector limit’, one gains access to a large
number of tools that allow one to rigorously establish a framework under which word2vec
can enjoy provable guarantees, such as in \cite{singer2017spectral,belkin2007convergence}. 

\section{Results}

We now state our main results. In \S 3.1 we establish that the energy functional $L(w,v)$ has a nice asymptotic expansion around $(v,w) = (0,0) \in \mathbb{R}^n \times \mathbb{R}^n$ and corresponds naturally to a spectral method in that regime. Naturally, such an asymptotic expansion is only feasible if one has some control over the size of the entries of the extremizer. We establish in \S 3.2 that the vectors maximizing the functional are not too large. The results in \S 3.2 are closely matched by numerical results: in particular, we observe that $\|w\| \sim \sqrt{n}$ in practice, a logarithmic factor smaller than our upper bound. The proofs are given in \S 4 and explicit numerical examples are shown in \S 5. In \S 6 we show empirically that the relation between word2vec and the spectral approach holds also for embedding in more than one dimension.
\subsection{First order approximation for small data.} The main idea is simple: we make an ansatz assuming that the optimal vectors are roughly of size $\|w\|,\|v\|\sim 1 $. If we assume that the vectors $w,v$ are fairly `typical' vectors of size $\sim 1$, 
then each entry is expected to scale approximately as $\sim n^{-1/2}$. 
Our main observation is that this regime is governed by a regularized spectral method. Before stating our theorem, let $ \lesssim $ denote the inequality up to universal multiplicative constants.

\begin{theorem}[Spectral Expansion]\label{thm:second_approx}
If $\|v\|_{\infty}, \|w\|_{\infty} \lesssim n^{-1/2}$, then
$$ L(w, v) = \left\langle w, Pv \right\rangle - \frac{1}{n} \left(\sum_{i=1}^{n}{w_i} \right) \left(  \sum_{j=1}^{n}{v_j}  \right) - \frac{1}{n} \sum_{i,j = 1}^{n}{\frac{w_i^2 v_j^2}{2}}  - n \log{n} + \mathcal{O}(n^{-1}).$$
\end{theorem}
Naturally, since we are interested in maximizing this quantity, the constant factor $n \log{n}$ plays no role. The leading terms can be rewritten as
$$ \left\langle w, Pv \right\rangle - \frac{1}{n} \left(\sum_{i=1}^{n}{w_i} \right) \left(  \sum_{j=1}^{n}{v_j}  \right)  = \left\langle w, \left( P - \frac{1}{n} \textbf{1} \right) v \right\rangle,$$
where $\textbf{1}$ is the matrix all of whose entries are 1. This suggests that the optimal $v,w$ maximizing the quantity should simply be the singular vectors associated
to the matrix $ P - \frac{1}{n} \textbf{1}$.
The full expansion has a quadratic term that serves as an additional regularizer.
The symmetric case (with ansatz $v = w$) is particularly simple, since we have
$$ L(w) = \left\langle w, Pw \right\rangle - \frac{1}{n} \left(\sum_{i=1}^{n}{w_i} \right)^2 - \frac{\|w\|^4}{2n}  -n \log{n} + \mathcal{O}(n^{-1}).$$
Assuming $P$ is similar to a symmetric matrix, the optimal $w$ should be well described by the leading eigenvector of $\left( P - \frac{1}{n} \textbf{1} \right)$ with 
an additional regularization term ensuring that $\|w\|$ is not too large. We consider this simple insight to
be the main contribution of this paper, since it explains succinctly why an algorithm like word2vec has a chance to be successful. We also give a
large number of numerical examples showing that in many cases the result obtained by word2vec is extremely similar to what we obtain from
the associated spectral method.

\vspace{1.5em}

\subsection{Optimal vectors are not too large} 
Another basic question is as follows: how large is the norm of the vector(s) maximizing the energy function?  This is of obvious importance in practice, however, as seen in Theorem 3.1, it
also has some theoretical relevance: if $w$ has large entries, then clearly one cannot hope to capture the exponential nonlinearity with a polynomial expansion. Assuming $\|P\| \leq 1$, the global maximizer $w^*$ of the second-order approximation
\begin{equation}\label{eq:sec_approximation} 
L_2(w) = \left\langle w, Pw \right\rangle - \frac{1}{n} \left(\sum_{i=1}^{n}{w_i} \right)^2 - \frac{\|w\|^4}{2n}  - n \log{n},
\end{equation}
satisfies
$$ \|w^*\| \leq  \sqrt{2n}.$$
This can be seen as follows: if $\|P\| \leq 1$, then $\left\langle w, Pw \right\rangle \leq \|w\|^2$. Plugging in $w=0$ shows that the
maximal energy is at least size $-n\log{n}$. For any vector exceeding $\sqrt{2n}$ in size, we see that the energy is less than that establishing the bound.
We obtain similar boundedness properties for the fully nonlinear problem for a fairly general class of matrices.
\begin{theorem}[Generic Boundedness.] Let $P \in \mathbb{R}^{n \times n}$ satisfy $\|P\| < 1$. Then
$$ w = \arg\max_{w} ~~  \left\langle w, Pw\right\rangle - \sum_{i=1}^{n} \log\left( \sum_{j=1}^{n} \exp(w_i w_j)\right),$$
satisfies
$$ \| w\|^2 \leq  \frac{n \log{n}}{1-\|P\|}.$$
\end{theorem}
While we do not claim that this bound is sharp, however it does nicely illustrate that the solutions of the optimization problem must be bounded. Moreover,
if they are bounded, then so are their entries; more precisely, $\|w\|^2 \lesssim n$ implies that, for `flat' vectors, the typical entry is of size $\lesssim 1$ and thus
firmly within the approximations that can be reached by a Taylor expansion. 
It is clear that a condition such as $\|P\| < 1$ is required for boundedness of the solutions. This can be observed by considering the row-stochastic matrix
$$ P = \begin{pmatrix}  1-\varepsilon & \varepsilon \\ \varepsilon & 1-\varepsilon \end{pmatrix}.$$
Writing $w = (w_1, w_2)$, we observe that the arising functional is quite nonlinear even in this simple case. However, it is fairly easy to understand the behavior of the gradient ascent method on the $w_1-$axis since
 \begin{eqnarray*}
\frac{\partial}{\partial w_1} L(w_1, w_2) \big|_{w_2=0} = 2 w_1 \left( 1 - \varepsilon -  \frac{e^{w_1^2}}{1 + e^{w_1^2}}  \right),   
 \end{eqnarray*}
is monotonically increasing until $w_1 \sim \pm \sqrt{\log{\varepsilon^{-1}}}$. Therefore it is, a priori, unbounded since $\varepsilon$ can be arbitrarily close to 0.

In practice, one often uses word2vec for matrices whose spectral norm is $\|P\|=1$ and which have the additional property of being row-stochastic. 
We also observe empirically
that the global optimizer $w^*$ has a mean value close to 0 (the expansion in Theorem 3.1 suggests why this would be the case). We achieve a similar boundedness theorem in which the only relevant operator norm is that of the operator restricted to the subspace of vectors having mean 0.

\begin{theorem}
	[Boundedness for row-stochastic matrices]
	\label{thm:upper_bound}
	 Let $P \in \mathbb{R}^{n \times n}$ be a row-stochastic matrix and let 
	 $$ P_S: \left\{w \in \mathbb{R}^n: w_1 + \dots + w_n = 0\right\} \rightarrow \mathbb{R}^n,$$
	 denote the restriction of $P$ to that subspace
	  and suppose that $\|P_S\| < 1$. Let
$$ w = \arg\max_{w} ~~  \left\langle w, Pw\right\rangle - \sum_{i=1}^{n} \log\left( \sum_{j=1}^{n} \exp(w_i w_j)\right).$$
If $w$ has a mean value sufficiently close to 0,
$$ \left|\left\langle w, \frac{\textbf{1}}{\sqrt{n}} \right\rangle\right| \leq \frac{1-\|P_S\|}{3}\|w\|,$$
where $\textbf{1} = (1,1,1,\dots, 1)$, then
$$ \|w\|^2 \leq \frac{2n \log{n}}{1 - \|P_S\|}.$$
\end{theorem}

The $2 \times 2$ matrix given above, illustrates that some restrictions are necessary, in order to obtain a nicely bounded gradient ascent. 
There is some freedom in the choice of the constants in Theorem 3.3.
Numerical
experiments show that the results are not merely theoretical: extremizing vectors tend to have a mean value sufficiently close to 0 for the theorem
to be applicable. 
\vspace{1.5em}

\subsection{Outlook}
Summarizing, our main arguments are as follows:
\begin{enumerate}
\item The energy landscape of the word2vec functional is well approximated by a spectral method (or regularized spectral method) as long as the entries of
the vector are uniformly bounded. In any compact interval around 0, the behavior of the exponential function can be appropriately
approximated by a Taylor expansion of sufficiently high degree.

\item There are bounds that suggests that the energy of the embedding vector scale as $\sqrt{n\log n}$; this means that, for `flat' vectors, the individual entries grow at most like $\sqrt{\log{n}}$. Presumably this is an artifact of the proof.

\item Finally, we present examples in \S 4 showing that in many cases the embedding obtained by maximizing the word2vec functional are indeed accurately predicted by the second order approximation.
\end{enumerate}
This suggests various interesting lines of research: it would be nice to have refined versions of Theorem 3.2 and Theorem 3.3 (an immediate goal being
the removal of the logarithmic dependence and perhaps even pointwise bounds on the entries of $w$). Numerical experiments indicate that Theorem 3.2 and Theorem 3.3 are at
most a logarithmic factor away from being optimal. A second natural avenue of research proposed by our paper is to differentiate the behavior of
word2vec and that of the associated spectral method: are the results of word2vec (being intrinsically nonlinear) truly different from the behavior of the
spectral method (arising as its linearization)? Or, put differently, is the nonlinear aspect of word2vec that is \textit{not} being captured by the spectral method
helpful for embedding?

\section{Proofs}
\begin{proof}[Proof of Theorem 3.1]
We recall our assumption of $\|w\|_{\infty} \lesssim n^{-1/2}$ and $\|v\|_{\infty} \lesssim n^{-1/2}$ (where the implicit constant affects all subsequent constants).
We remark that the subsequent arguments could also be carried out for any $\|w\|_{\infty}, \|v\|_{\infty} \lesssim n^{-\varepsilon}$ at the cost of different error terms;
the arguments fail to be rigorous as soon as $\|w\|_{\infty} \sim 1$, since then, a priori, all terms in the Taylor expansion of $e^x$ could be of roughly the
same size. We start with the Taylor expansion
\begin{align*}
\sum_{j=1}^n e^{w_i v_j} &= \sum_{j=1}^n  \left( 1 + w_i v_j + \frac{ w_i^2 v_j^2}{2} +   \mathcal{O}\left( n^{-3} \right)\right) \\
&= n +  \sum_{j=1}^n  \left(w_i v_j + \frac{ w_i^2 v_j^2}{2}  \right)  +  \mathcal{O}\left( n^{-2} \right).
\end{align*}
In particular, we note that
$$ \left| \sum_{j=1}^n  \left(w_i v_j + \frac{ w_i^2 v_j^2}{2}  \right)\right| \lesssim 1.$$
We use the series expansion
$$ \log{(n + x)} = \log{n} + \frac{x}{n} - \frac{x^2}{2n^2} + \mathcal{O}\left(\frac{|x|^3}{n^3}\right)$$
to obtain
\begin{align*}
\log \left(\sum_{j=1}^n e^{w_i v_j} \right) &= \log{n} + \frac{1}{n}  \sum_{j=1}^n \left(w_i v_j + \frac{ w_i^2 v_j^2}{2}\right)  \\
&- \frac{1}{2n^2} \left( \sum_{j=1}^n  w_i v_j + \frac{ w_i^2 v_j^2}{2}   \right)^2 + \mathcal{O}(n^{-3}).
\end{align*}
Here, the second sum can be somewhat simplified since 
\begin{align*}
 \frac{1}{2n^2} \left( \sum_{j=1}^n  w_i v_j + \frac{ w_i^2 v_j^2}{2}   \right)^2 &=  \frac{1}{2n^2} \left( \sum_{j=1}^n  \left(w_i v_j + \mathcal{O}(n^{-2}) \right)  \right)^2\\
  &=  \frac{1}{2n^2} \left(\mathcal{O}(n^{-1}) +  \sum_{j=1}^n w_i v_j  \right)^2 \\
  &=   \frac{1}{2n^2} \left( \sum_{j=1}^n w_i v_j    \right)^2 + \mathcal{O}(n^{-3})\\
  &=   \frac{w_i^2}{2n^2} \left( \sum_{j=1}^n  v_j    \right)^2 + \mathcal{O}(n^{-3})\\
 \end{align*}
Altogether, we obtain that
\begin{align*}
\sum_{i=1}^n \log \left(\sum_{j=1}^n e^{w_i v_j} \right) &= \sum_{i=1}^{n} \left( \log{n} + \frac{1}{n}  \sum_{j=1}^n \left(w_i v_j + \frac{ w_i^2 v_j^2}{2}\right) -  \frac{w_i^2}{2n^2} \left( \sum_{j=1}^n  v_j    \right)^2 + \mathcal{O}(n^{-3}) \right)\\
&= n \log{n} + \frac{1}{n} \sum_{i,j = 1}^{n}{w_i v_j} + \frac{1}{n} \sum_{i,j = 1}^{n}{\frac{w_i^2 v_j^2}{2}} \\
&- \frac{1}{n^2} \left( \sum_{i=1}^{n}{ \frac{w_i^2}{2}} \right) \left(\sum_{j=1}^{n}{ v_j} \right)^2 + \mathcal{O}(n^{-2}).
\end{align*}
Since  $\|w\|_{\infty}, \|v\|_{\infty} \lesssim n^{-1/2}$, we have
$$    \frac{1}{n^2} \left( \sum_{i=1}^{n}{ \frac{w_i^2}{2}} \right) \left(\sum_{j=1}^{n}{ v_j} \right)^2 \lesssim n^{-1}$$
and have justified the desired expansion.
\end{proof}

\begin{proof}[Proof of Theorem 3.2]
Setting $w = 0$ results in the energy
$$ L(w) = -n \log{n}.$$
Now, let $w$ be a global maximizer. We obtain
\begin{align*}
 - n \log{n} &\leq \left\langle w, Pw \right\rangle - \sum_{i=1}^n \log \left(\sum_{j=1}^n e^{w_i w_j} \right) \\
&\leq  \|P\| \|w\|^2 - \sum_{i=1}^n \log \left( e^{w_i^2} \right) \leq (\|P\| - 1)\|w\|^2
\end{align*}
which is the desired result.
\end{proof}

\begin{proof}[Proof of Theorem 3.3]
We expand the vector $w$ into a multiple of the constant vector of norm 1, the vector
$$ \frac{\textbf{1}}{\sqrt{n}} = \left( \frac{1}{\sqrt{n}}, \frac{1}{\sqrt{n}}, \dots, \frac{1}{\sqrt{n}} \right),$$
and the orthogonal complement via
$$ w  = \left\langle w,  \frac{\textbf{1}}{\sqrt{n}}  \right\rangle  \frac{\textbf{1}}{\sqrt{n}}  + \left(w -  \left\langle w,  \frac{\textbf{1}}{\sqrt{n}}  \right\rangle  \frac{\textbf{1}}{\sqrt{n}}  \right),$$
which we abbreviate as $w = \tilde w + (w - \tilde w)$. We expand, 
\begin{align*}
\left\langle w, Pw \right\rangle &= \left\langle \tilde w, P \tilde w \right\rangle + \left\langle \tilde w, P(w-\tilde w) \right\rangle + \left\langle w-\tilde w, P\tilde w \right\rangle + \left\langle w-\tilde w, P(w-\tilde w) \right\rangle.
\end{align*}
Since $P$ is row-stochastic, we have $P \tilde w = \tilde w$ and thus $ \left\langle \tilde w, P \tilde w \right\rangle  = \|\tilde w\|^2$. Moreover, we have
$$  \left\langle w-\tilde w, P\tilde w \right\rangle  =  \left\langle w-\tilde w, \tilde w \right\rangle = 0$$
since $w - \tilde w$ has mean value 0. We also observe, again because $w - \tilde w$ has mean value 0, that 
$$  \left\langle \tilde w, P(w-\tilde w) \right\rangle  =  \left\langle \tilde w, P_S(w-\tilde w) \right\rangle.$$
Collecting all these estimates, we obtain
$$ \frac{ \left\langle w, Pw \right\rangle}{\|w\|^2} \leq  \frac{\|\tilde w\|^2}{\|w\|^2} +  \frac{\|\tilde w \|}{\|w\|}  \frac{ \|w - \tilde w\|}{\|w\|} \|P_S\| +  \frac{ \|w - \tilde w\|^2}{\|w\|^2} \|P_S\|.$$
We also recall the Pythagorean theorem,
$$ \|\tilde w\|^2 + \|w - \tilde w\|^2 = \|w\|^2.$$
Abbreviating $x=\|\tilde w\|/\|w\|$, we can abbreviate our upper bound as
$$ \frac{ \left\langle w, Pw \right\rangle}{\|w\|^2} \leq x^2 + x\sqrt{1-x^2} \|P_S\| + (1-x^2)\|P_S\|.$$
The function,
$$ x \rightarrow  x\sqrt{1-x^2} + (1-x^2)$$
is monotonically increasing on $[0,1/3]$. Thus, assuming that
$$  x = \frac{\|\tilde w\|}{\|w\|}  \leq \frac{1-\|P_S\|}{3},$$
we get, after some elementary computation,
\begin{align*}
 \frac{ \left\langle w, Pw \right\rangle}{\|w\|^2} &\leq  \left(\frac{1-\|P_S\|}{9} \right)^2 + \frac{1-\|P_S\|}{9} \sqrt{1-\left(\frac{1-\|P_S\|}{9} \right)^2} \|P_S\| \\
 &+ \left(1-\left(\frac{1-\|P_S\|}{9} \right)^2\right)\|P_S\| \leq 0.2 + 0.8\|P_S\|.
 \end{align*}
However, we also recall from the proof of Theorem 3.2 that
$$ \sum_{i=1}^{n} -\log\left( \sum_{j=1}^{n}{ e^{w_i w_j}} \right) \leq - \|w\|^2.$$
Altogether, since the energy in the maximum has to exceed the energy in the origin, we have
\begin{align*}
-n \log{n} \leq \left\langle w, Pw \right\rangle -  \sum_{i=1}^{n} \log\left( \sum_{j=1}^{n}{ e^{w_i w_j}} \right) \leq \left( 0.2 + 0.8 \|P_S\| \right)\|w\|^2 - \|w\|^2
\end{align*}
and therefore,
$$ \|w\|^2 \leq \frac{2 n \log{n}}{1 - \|P_S\|}.$$
\end{proof}

\section{Examples}
\label{sec:Examples}
We validate our theoretical findings by comparing, for various datasets, the representation obtained by the following methods: (i) optimizing over the symmetric functional in  \eqref{eq:skip_gram_function}, (ii) optimizing over the spectral method suggested by Theorem   \ref{thm:second_approx} and (iii) computing the leading eigenvector of $P - \frac{1}{n} \textbf{1}$. 
We denote by $w$, $\hat w$ and $u$ be the three vectors obtained by (i)-(iii) respectively. 
The comparison is performed for two artificial datasets, two sets of images, a seismic dataset and a text corpus.
For the artificial, image and seismic data, the matrix $P$ is obtained by the following steps: we compute a pairwise kernel matrix 
\begin{eqnarray*}
K(x_i,x_j) =  \exp\bigg( -\frac{||x_i-x_j||^2}{\alpha}\bigg),
\end{eqnarray*}
where $\alpha$ is a scale parameter set as in \cite{lafon2006data} using a max-min scale. The max-min scale is set to
\begin{eqnarray} \label{eq:MaxMin}
\alpha= \underset{j}{\max} [ \underset{i,i\neq j}{\min} (||{x}_i-{x}_j||^2)],i,j=1,...n.
\end{eqnarray} This global scale guarantees that each point is connected to at least one other point. Alternatively, adaptive scales could be used as suggested in \cite{kernelscaling}. We then compute $P$ via
\begin{eqnarray*}
	P_{ij} = K_{ij}\big/\sum_{l=1}^N K_{il}.
\end{eqnarray*}
The matrix $P$ can be interpreted as a random walk over the data points, (see for example \cite{coifman2006diffusion}). Theorem \ref{thm:second_approx} holds for any matrix $P$ that is similar to a symmetric matrix, here we use the common construction from \cite{coifman2006diffusion}, but our results hold for other variations as well.
To support our approximation in Theorem  \ref{thm:second_approx}, we compute the correlation coefficient between $u$ and $\hat w$ by
\begin{eqnarray*}
\rho(u,\hat w)=\frac{(u-\mu)^T(\hat w-\hat \mu)}{\|u-\mu\| \|\hat w- \hat \mu\|},
\end{eqnarray*} 
where $\mu$ and $\hat \mu$ are the means of $u$ and $\hat w$ respectively.
A similar measure is done for $ w$ and $u$.
In addition, we illustrate that the norm $\|w\|$ is comparable to $\sqrt{n}$, which supports the upper bound in Theorem \ref{thm:upper_bound}.
\subsection{Noisy Circle}
\label{subsec:noisy_circle}
Here, the elements $\{x_1,...,x_{200} \in \mathbb{ R}^2 \}$ are generated by adding Gaussian noise with mean $0$ and $\sigma^2=0.1$ to a unit circle (see the left panel of Figure \ref{fig:circle}). The right panel shows the extracted representations $w$, $\hat{w}$ along with the leading eigenvector $u$ scaled by $\sqrt{\lambda n}$ where $\lambda$ is the corresponding eigenvalue. The correlation coefficients $\rho(w,u)$ and $\rho(\hat{w},u)$ are equal to $0.98$, $0.99$ respectively. 

\subsection{Binary MNIST}
Next, we use a set of $300$ images of the digits $3$ and $4$ from the MNIST dataset \cite{mnist}. Two examples from each category are presented in the left panel of Figure \ref{fig:mnist}. Here, the extracted representations $w$ and $\hat{w}$ match the values of the scaled eigenvector $u$ (see right panel of Figure \ref{fig:mnist}). The correlation coefficients $\rho(w,u)$ and $\rho(\hat w,u)$ are both higher than $0.999$. 

\subsection{COIL100}
In this example, we use images from Columbia Object Image Library (COIL100) \cite{coil100}. Our dataset contains $21$ images of a cat  captured at several pose intervals of $5$ degrees (see left panel of Figure \ref{fig:coil100}). We extract the embedding $w$ and $\hat{w}$ and reorder them based on the true angle of the cat at every image. In the right panel, we present the values of the reordered representations $w$, $\hat{w}$ and $u$ overlayed with the corresponding objects. The values of all representations are strongly correlated with the angle of the object. Moreover, the correlation coefficients $\rho(w,u)$ and $\rho(\hat w,u)$, are $0.97$ and $0.99$ respectively. 

\subsection{Seismic Data}
Seismic recordings could be useful for identifying properties of geophysical events. We use a dataset collected in Israel and Jordan, described in \cite{seismic}. The data consists of $1632$ seismic recordings of earthquakes and explosions from quarries. Each recording is described by a sonogram with $13$ frequency bins, and $89$ time bins \cite{joswig} (see the left panel of Figure \ref{fig:sonograms}). Events could be categorized into $5$ groups using manual annotations of their origin. We flatten each sonogram into a vector, and extract embeddings $w$, $\hat{w}$, and $u$. In the right panel of this figure, we show the extracted representations of all events. We use dashed lines to annotate the different categories and sort the values within each category based on $u$. The coefficient $\rho(w,v)$ is equal to $0.89$, and $\rho(\hat w,v)=1$.

\subsection{Text Data}
As a final evaluation we use a corpus of words from the book ``Alice in Wonderland'' as processed in \cite{alicetext}. To define a co-occurrence matrix, we scan the sentences using a window size covering $5$ neighbors before and after each word. We subsample the top $1000$ words in terms of occurrences in the book. The matrix $P$ is then defined by normalizing the co-occurrence matrix. In Figure \ref{fig:alice} we present centered and normalized versions of the representations $w$, $\hat{w}$ and the leading left singular vector $v$ of $P- \frac{1}{n} \textbf{1}$. The coefficient $\rho(w,v)$ is equal to $0.77$, and $\rho(\hat w,v)=1$.
\section{Multi-dimensional Embedding}
\label{sec:experiment high dim}
In Section \ref{eq:sec_approximation} we have demonstrated that under certain assumptions, the maximizer of the energy functional in \eqref{eq:symmetric_loss} is governed by a regularized spectral method. For simplicity, we have restricted our analysis to a one dimensional symmetric representations, i.e.
$w=v\in R^N$. Here, we demonstrate empirically that this result holds even when embedding $n$ elements $x_1,\ldots,x_n$ in higher dimensions.

Let $w_i \in R^d$ be the embedding vector associated with $x_i$, where $d\ll n$ is the embedding dimension. The symmetric word2vec functional is given by
\begin{equation}\label{eq:high-dim}
    L(W) = \text{Tr}(W ^T P W) -\sum_{i=1}^n  \log \Bigg(  \sum_{j=1}^n \exp(w^T_i w_j) \Bigg),
\end{equation}
where $W=[w_1,\ldots,w_n]^T \in \mathbb{R}^{n\times d}$. A similar derivation to the one presented in Theorem \ref{thm:second_approx} (the one dimensional case) yields the following approximation of \eqref{eq:high-dim}, 
\begin{eqnarray}\label{eq:approx_highdim}
L_2 (W) = \text{Tr}\big( W^T (P  - \frac{1}{n}\V{1} \big)   W) - \frac{\|W^TW\|_F^2}{2n} -n\log n.  
\end{eqnarray}
Note that both the symmetric functional in \eqref{eq:high-dim} and its approximation in  \eqref{eq:approx_highdim} are invariant to multiplying $W$ with an orthogonal matrix. That is, $W$ and $WR$ produce the same value in both functionals, where $R\in O(d)$.


To understand how the maximizer of \eqref{eq:high-dim} is governed by a spectral approach, we perform the following comparison. (i) We obtain the optimizer $ W$ of \eqref{eq:high-dim} via gradient descent, and compute its left singular vectors, denoted $u_1,\ldots,u_d$.
(ii) We compute the right singular vectors of $P-\frac1n \bf 1$, denoted by $\psi_1,\ldots,\psi_d$. (iii) Compute the pairwise absolute correlation values $\rho(u_i, \psi_j) .$


We experiment on two datasets: (1) A collection of $5$ Gaussians, and (2) images of hand written digits from MNIST. The transition probability matrix $P$ was constructed as described in Section \ref{sec:Examples}.


\vspace{0.2 in}
\subsection{Data from  distinct Gaussians}
In this experiment we generate a total of $2500$ samples, consisting of five sets of size $500$. The samples in the $i$-th set are drawn independently according to a Gaussian distribution $\mathcal N (r\cdot i\cdot  \textbf 1, 2\cdot I)$, where  $\textbf{1}$ is a $10-$ dimensional all ones vector, and $r$ is scalar that controls the separation between the Gaussian centers. 

Figure \ref{fig:corr_gauss_high} shows the absolute correlation value of the pairwise correlation between $u_1,\ldots,u_4$ and $\psi_1,\ldots,\psi_4$ for $r = 8,9, \text{ and } 10$. The correlation between the result obtained via the word2vec functional \eqref{eq:high-dim} and the right singular vectors of $P-\frac1n \textbf{1}$ increase when the separation between the Gaussians is high. 


\vspace{0.1 in}

\subsection{Multi-class MNIST}
The data consists of $10,000$ samples from the MNIST hand-written dataset with $1,000$ images from each digit ($0-9$). We compute a $10-$dimensional word2vec embedding $W$ by optimizing \eqref{eq:high-dim}.

Figure \ref{fig:highd2} shows the absolute correlation between the $u_1,\ldots,u_{10}$ and $\psi_1,\ldots,\psi_{10}$. As evident from the correlation matrix, the results obtained by both methods span similar subspaces. 


\subsection{Figures}
\begin{figure}[H]
	\centering
	\includegraphics[width=0.95\textwidth]{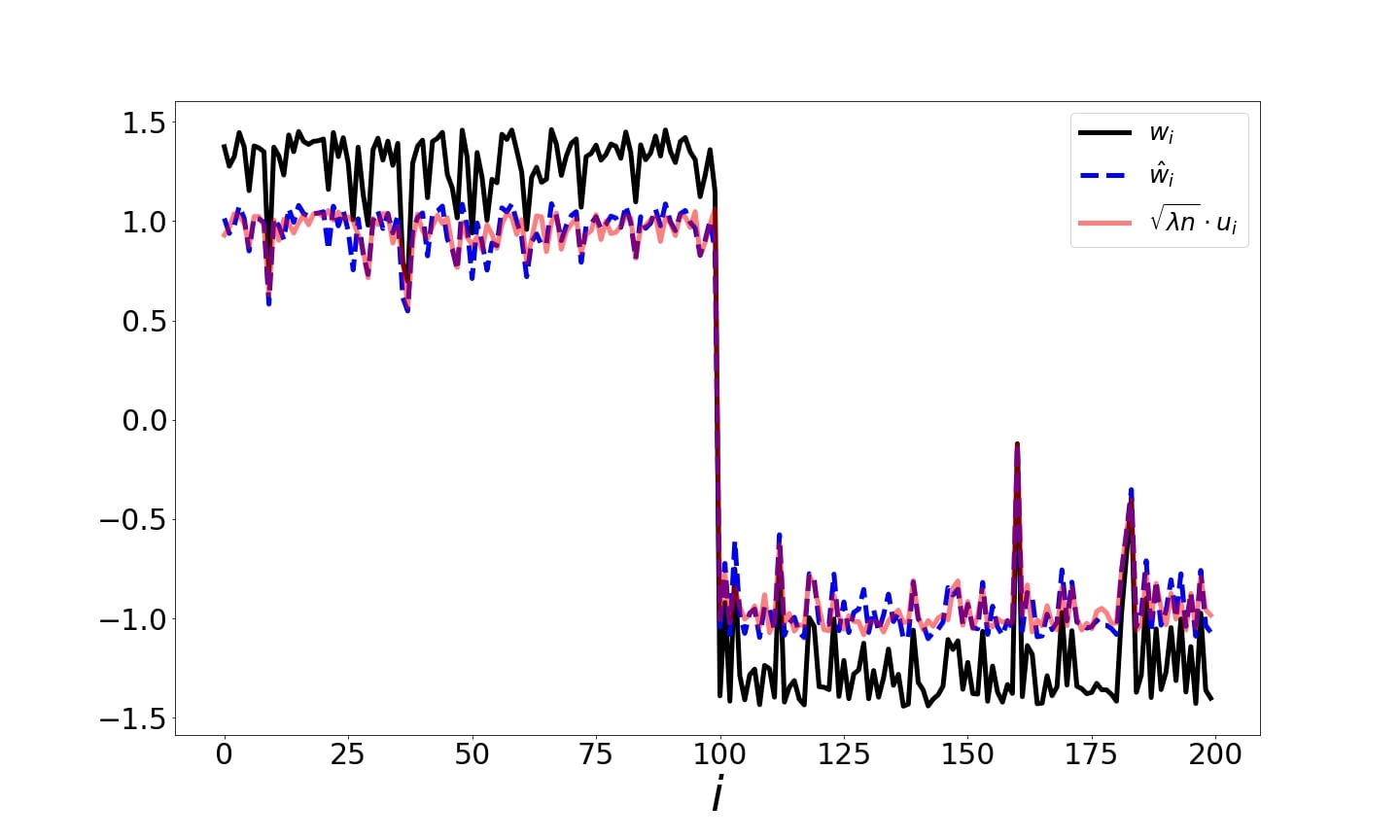}
	\caption{Illustration of point set drawn from two distinct Gaussian distributions. The result of maximizing over the word2vec functional (black) is closely tracked (up to scale) by the optimizer of the spectral method (blue) and the eigenvector (red). In Figure \ref{fig:hist}, we present a scatter plot comparing the values of $\hat{w}$ and $\sqrt{\lambda n} u$.}
	\label{fig:my_label}
\end{figure}

\begin{figure}[H]
	\centering
\includegraphics[width=1.\textwidth]{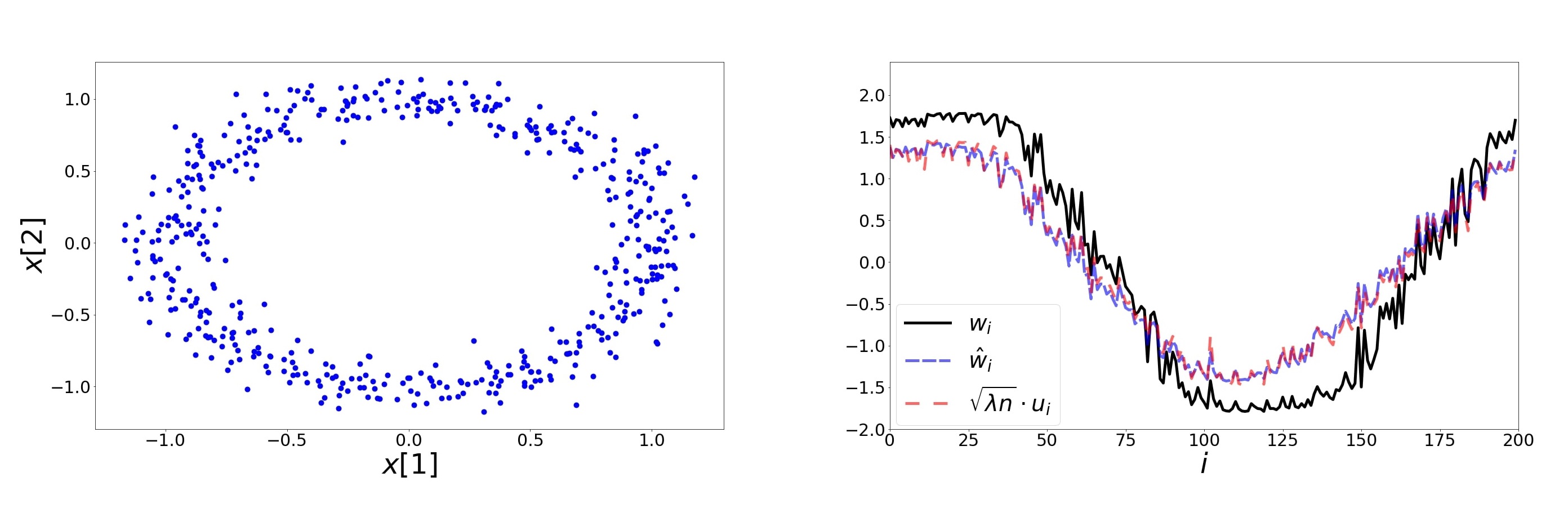}
	\caption{Left: $200$ elements on the noisy circle data set. Points are generated by adding noise drawn from a two dimensional Gaussian with zero mean and a variance of $0.1$. Right: The extracted representations based on the symmetric loss $w$, second order approximation $\hat w$ and leading eigenvector $u$. In Figure \ref{fig:hist}, we present a scatter plot comparing the values of $\hat{w}$ and $\sqrt{\lambda n} u$.}
	\label{fig:circle}
\end{figure}

\begin{figure}[H]
	\centering
	\includegraphics[width=0.45\textwidth]{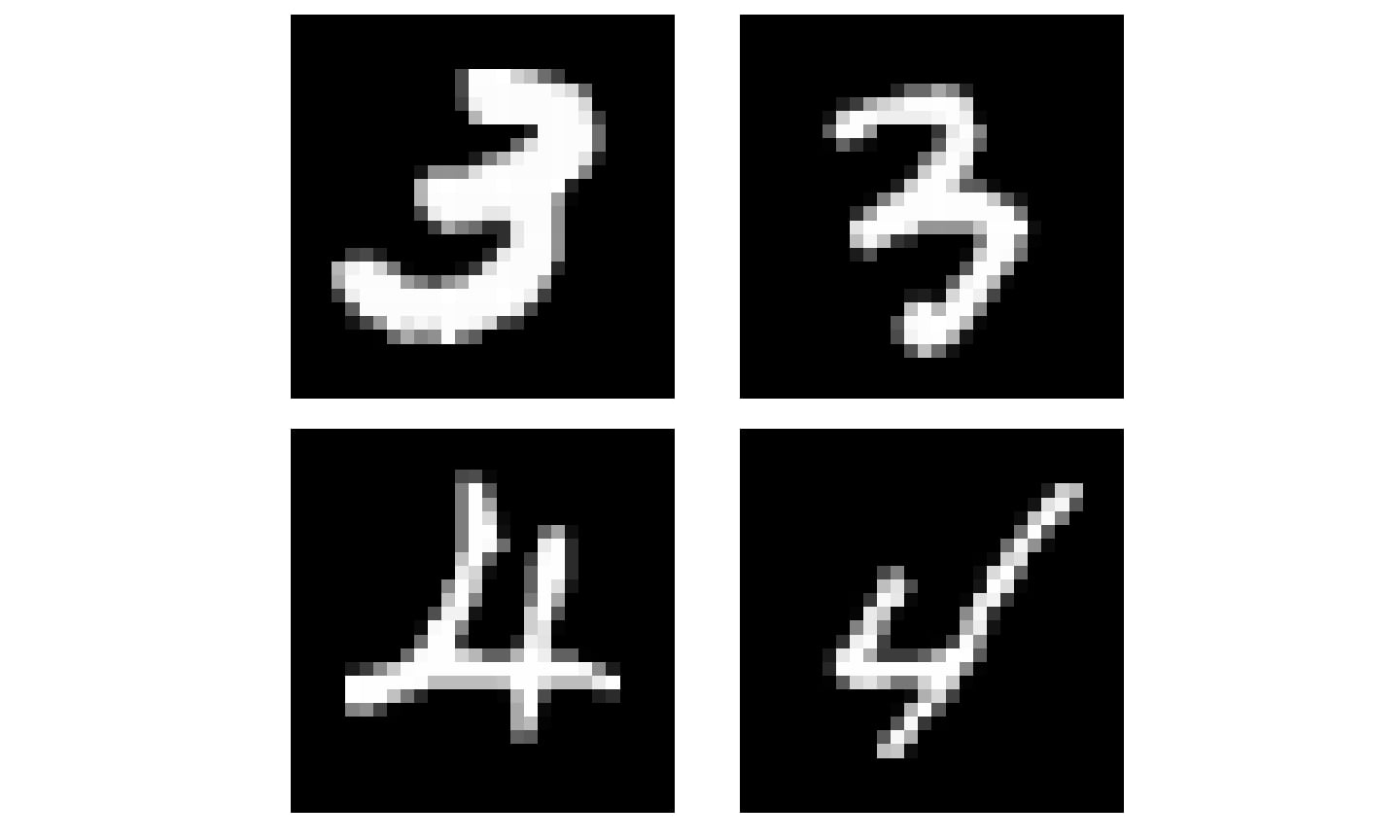}
	\includegraphics[width=0.5\textwidth]{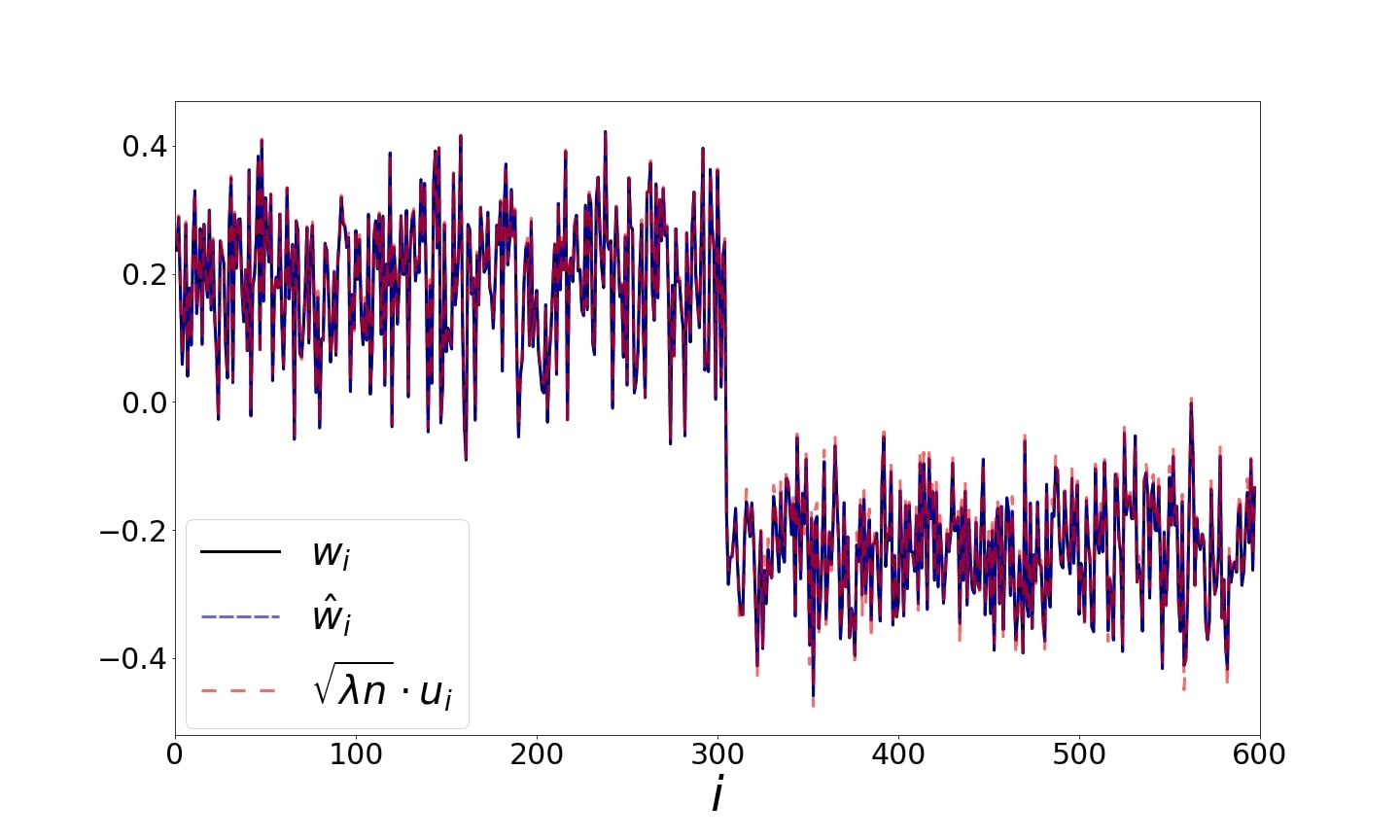}
	\caption{Left: handwritten digits from the MNIST dataset. Right: The extracted representations $w$, $\hat{w}$ and $\sqrt{\lambda n u}$, the leading eigenvector of $ P - \frac{1}{n} \textbf{1}$. In Figure \ref{fig:hist}, we present a scatter plot comparing the values of $\hat{w}$ and $\sqrt{\lambda n} u$. In Figure \ref{fig:hist}, we present a scatter plot comparing the values of $\hat{w}$ and $\sqrt{\lambda n} u$. }
	\label{fig:mnist}
\end{figure}

\begin{figure}[H]
	\centering
	\includegraphics[width=0.47\textwidth]{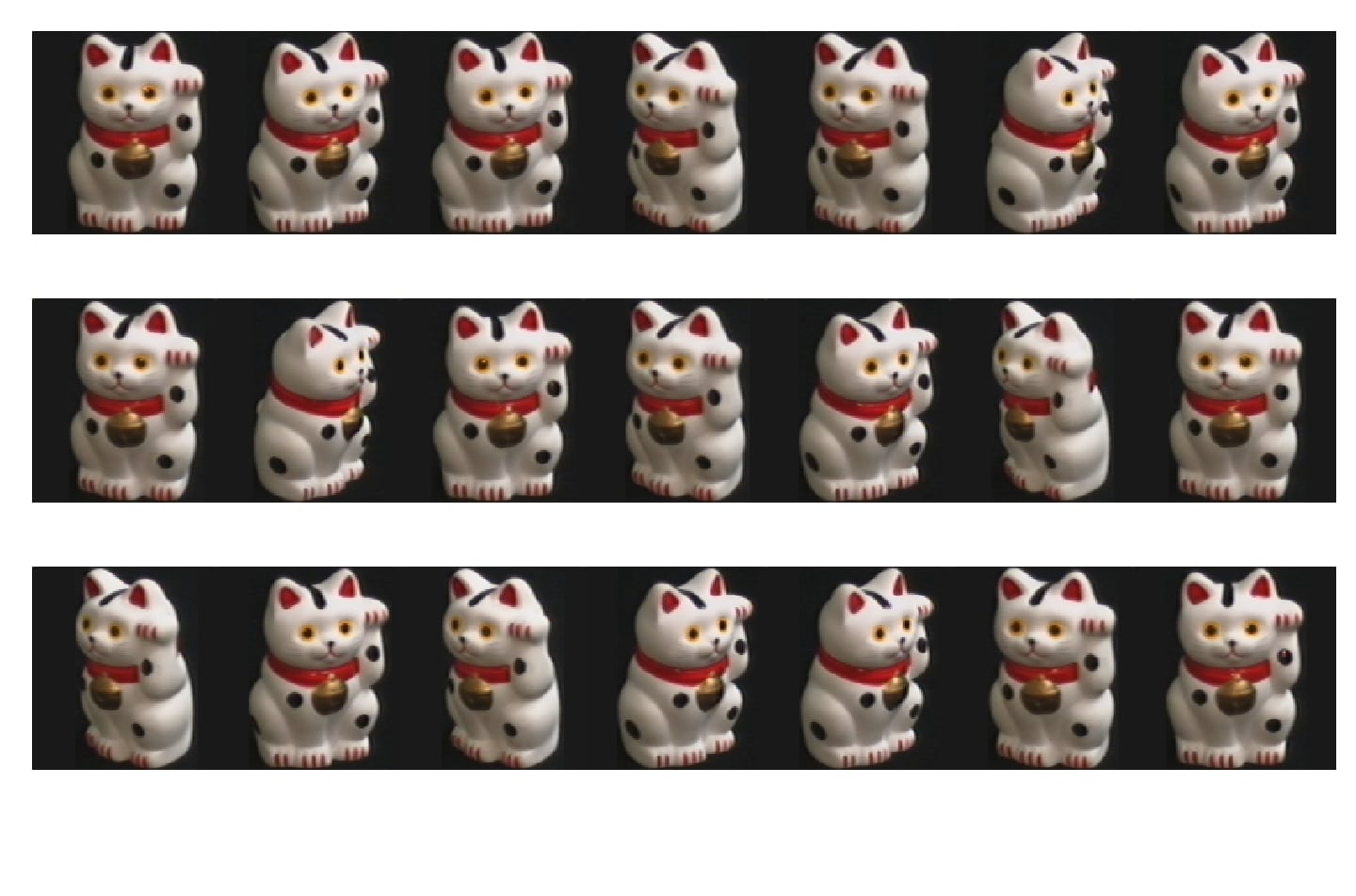}
	\includegraphics[width=0.52\textwidth]{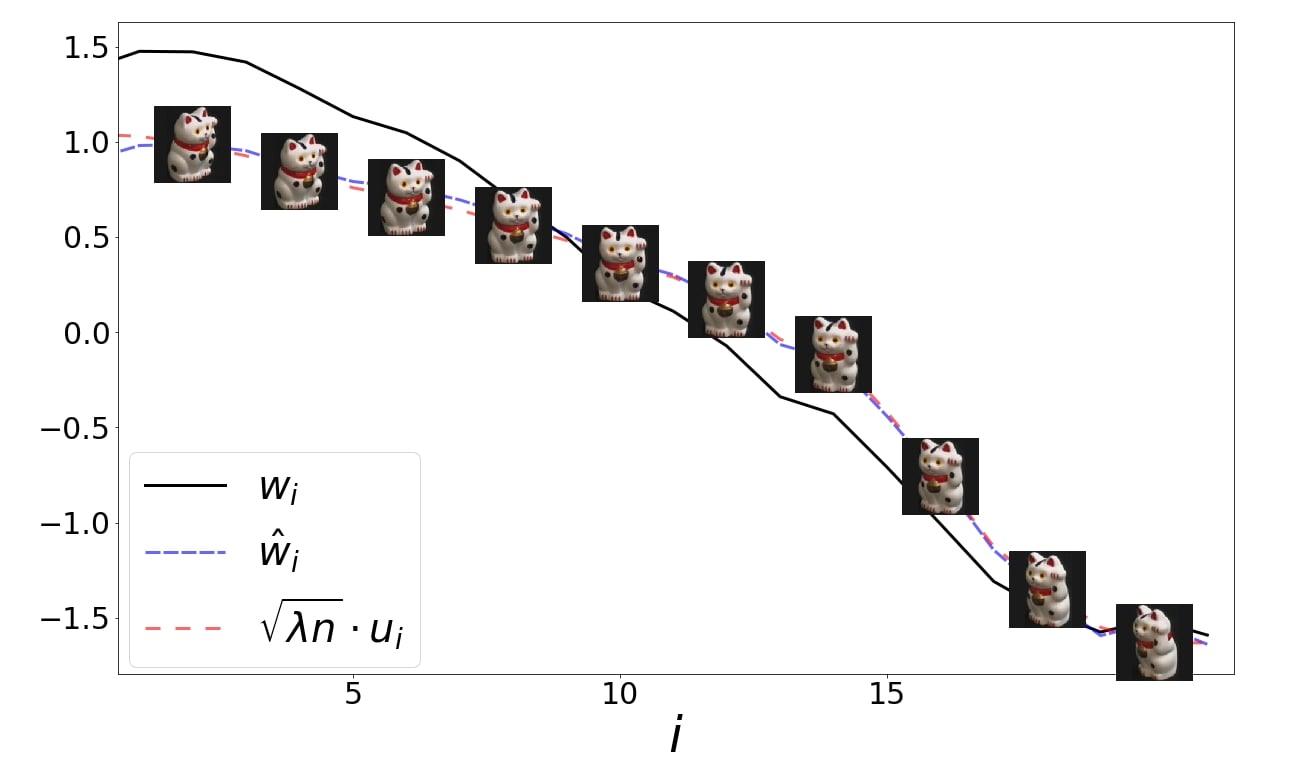}
	\caption{Left: $21$ samples from COIL100 dataset. The object is captured at several unorganized angles. 
	Right: 
	The sorted values of the representations $w$,$\hat{w}$ and $u$, along with the corresponding object. Here, the representation correlates with the angle of the object. In Figure \ref{fig:hist}, we present a scatter plot comparing the values of $\hat{w}$ and $\sqrt{\lambda n} u$.}
	\label{fig:coil100}
\end{figure}

\begin{figure}[H] 
	\centering
	\includegraphics[width=0.44\textwidth]{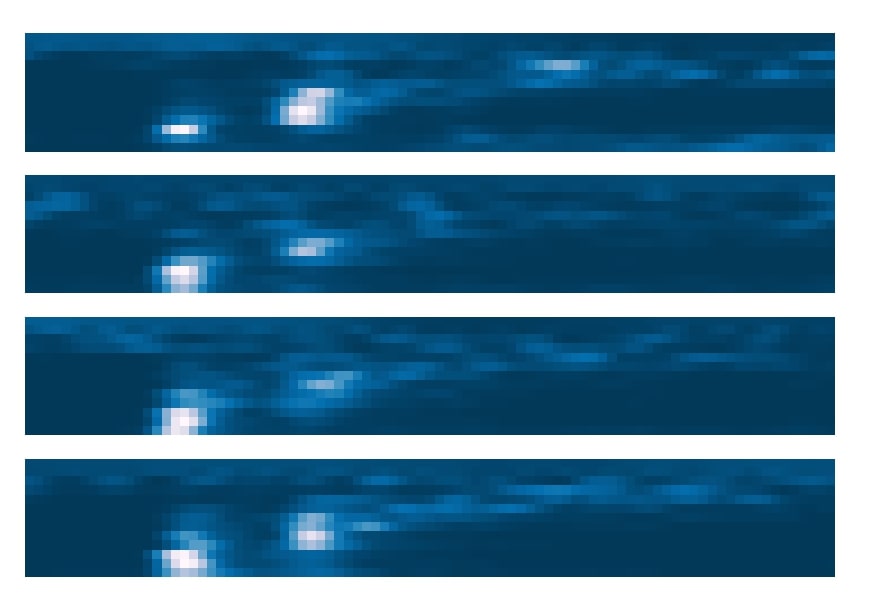}
	\includegraphics[width=0.55\textwidth]{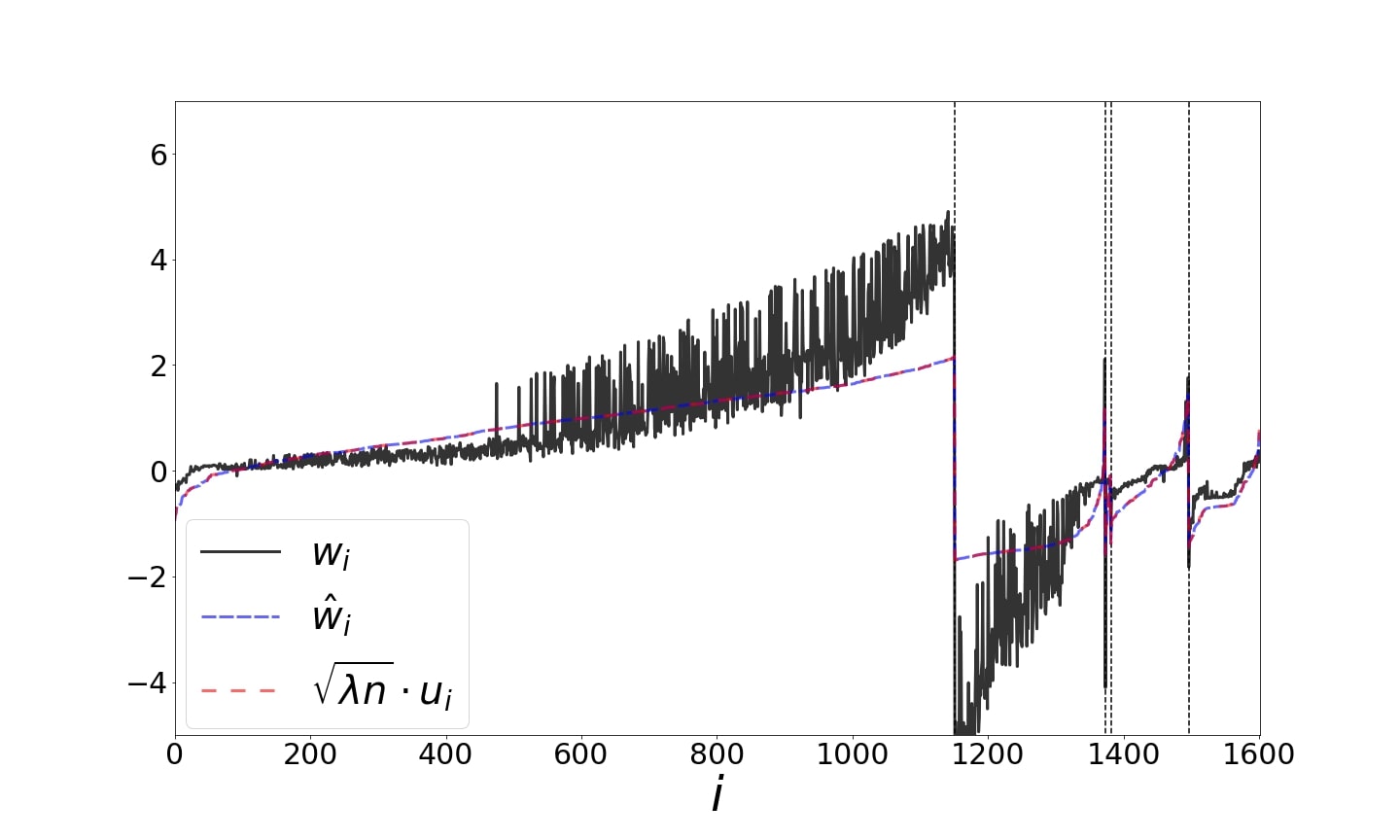}
	\caption{Left: $4$ samples from the sonogram dataset, of different event types.
	Right: 
	The values of the representations $w$,$\hat{w}$ and $u$. Dashed lines annotate the different categories of the events (based on event type and quarry location). Within each category the representations are ordered based on the value of $u$.}
	\label{fig:sonograms}
\end{figure}

\begin{figure}[H] 
	\centering
	\includegraphics[width=0.6\textwidth]{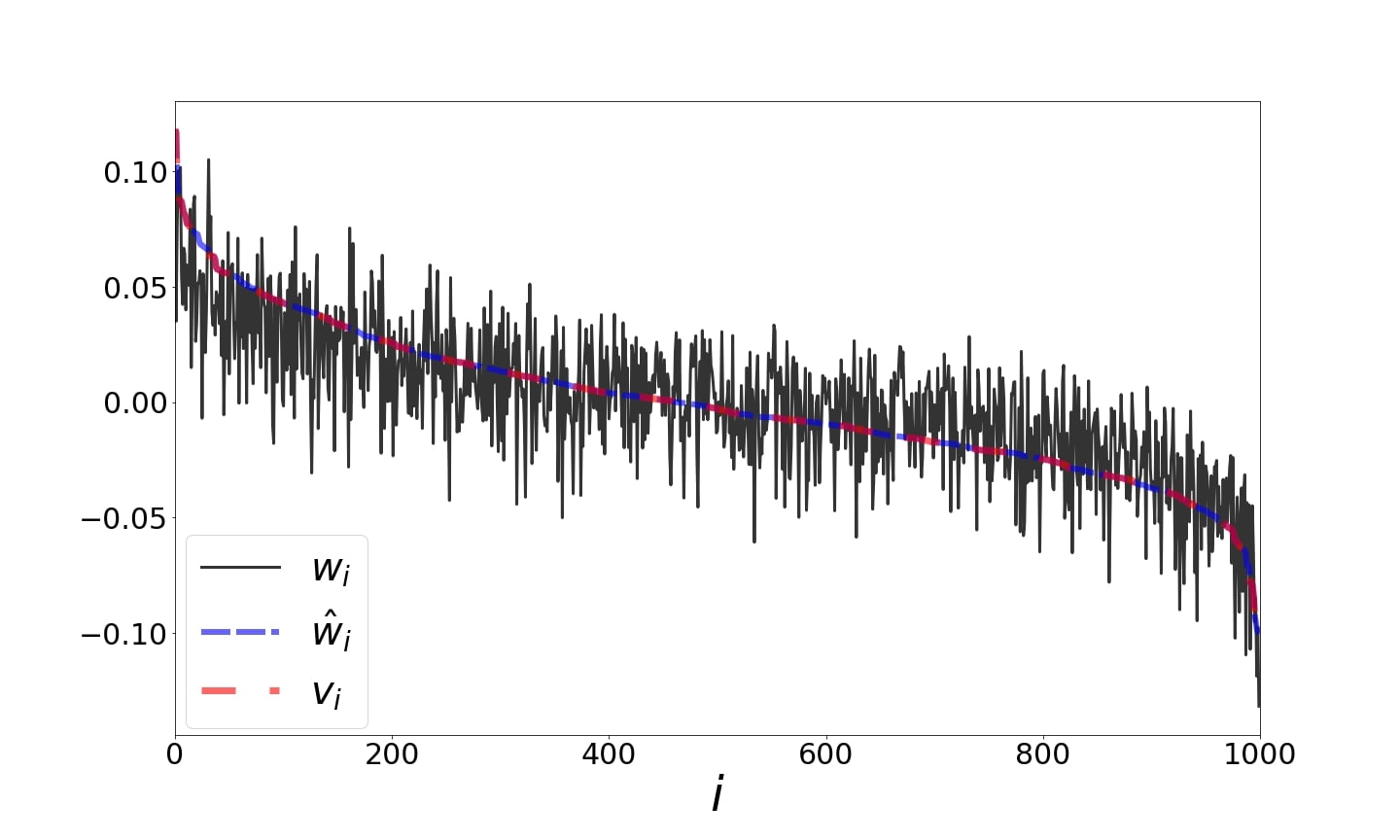}
	\caption{Word representation based on ``Alice in Wonderland''. The values of the representations $w$,$\hat{w}$ and $v$ are sorted based on the singular vector $v$. We normalized all representations to unit norm. }
	\label{fig:alice}
\end{figure}

\begin{figure}[H] 
	\centering
\includegraphics[width=1.\textwidth]{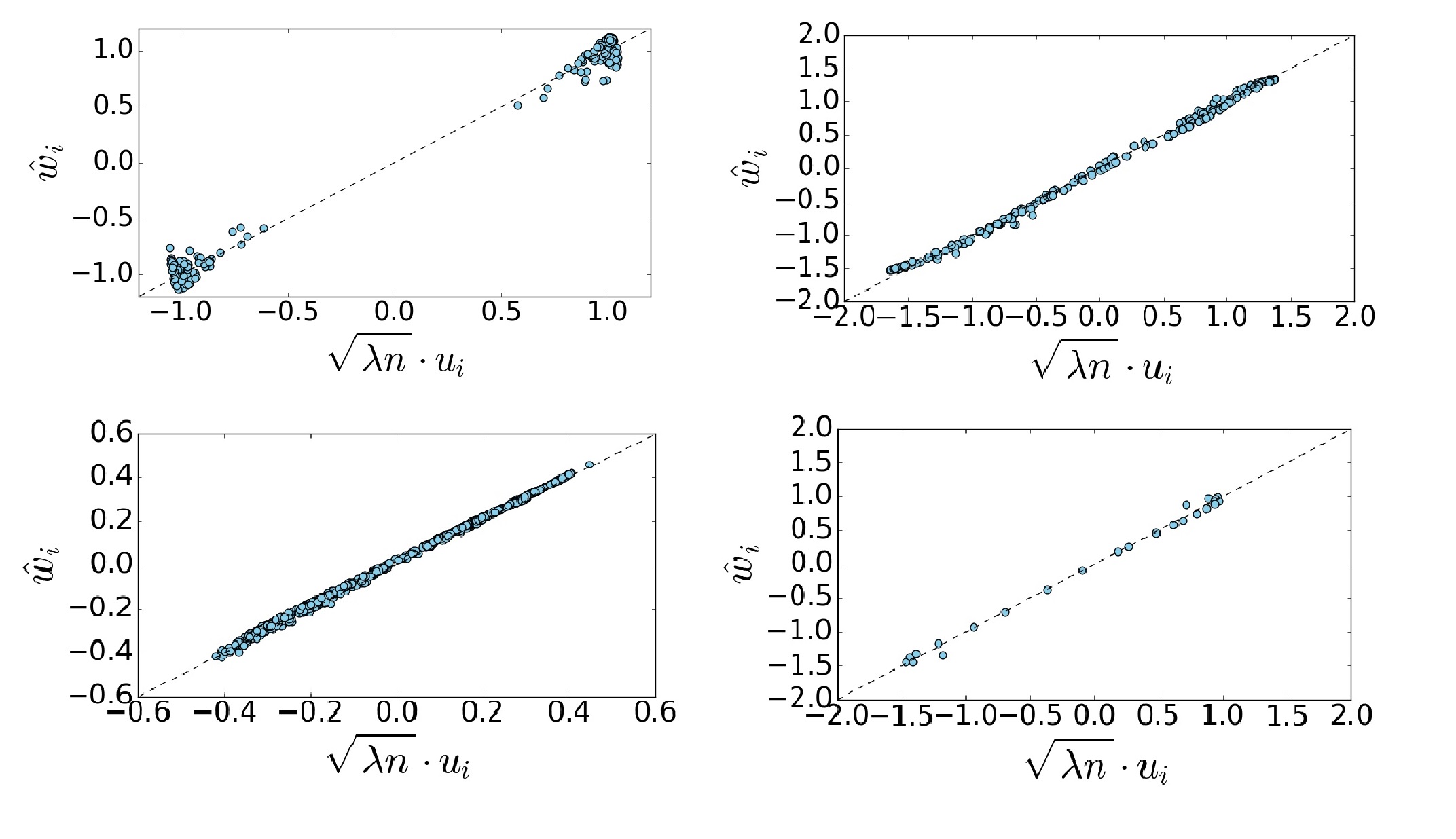}

	\caption{Scatter plots of the scaled eigenvector of $P-\frac{1}{n}\bf{1}$, denoted by $u$, and the minimizer of the approximated functional (in Eq. \ref{eq:sec_approximation}), denoted by $\hat{w}$. From top left to bottom right, results based on data from: two Gaussian clusters, a circle, binary MNIST images and COIL 100.
	}
	\label{fig:hist}
\end{figure}

\begin{figure}[H] 
	\centering
\includegraphics[width=0.3\textwidth]{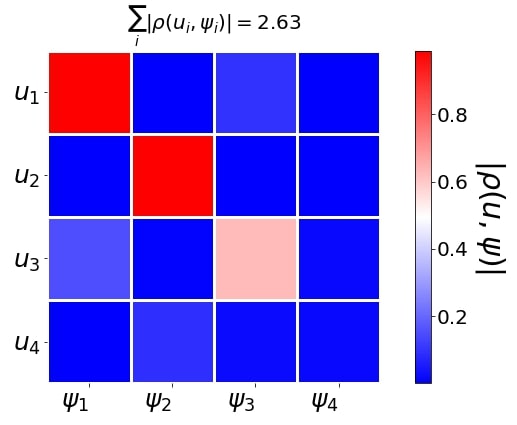}
\includegraphics[width=0.3\textwidth]{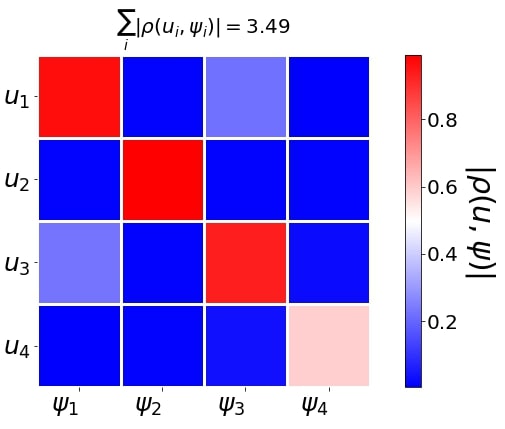}
\includegraphics[width=0.3\textwidth]{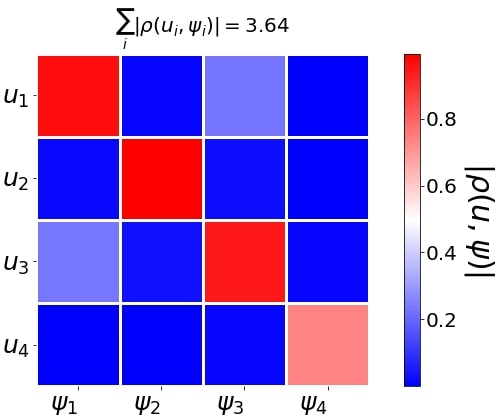}
	\caption{Word2vec embedding of a collection of Gaussians. We use $2500$ points generated according to five distinct Gaussians $\mathcal N (r\cdot i\cdot  \textbf 1, 2\cdot I)$, where $\textbf{1}$ is a $10-$ dimensional all ones vector, and $r$ is scalar that controls the separation between the Gaussian centers. The figure shows the absolute correlation between $u_1,\ldots,u_4$ and $\psi_1,\ldots,\psi_4$ for $r=8,9, \text{ and }10$. 
	For each value of $r$, we compute the sum of diagonal elements in the correlation matrix. 
	This results demonstrate the strong agreement between word2vec embedding and the spectral representation based on $P-\frac{1}{n} \textbf{1}$.
	}
	\label{fig:corr_gauss_high}
\end{figure}

\begin{figure}[H]
	\centering
	\includegraphics[width=0.45\textwidth]{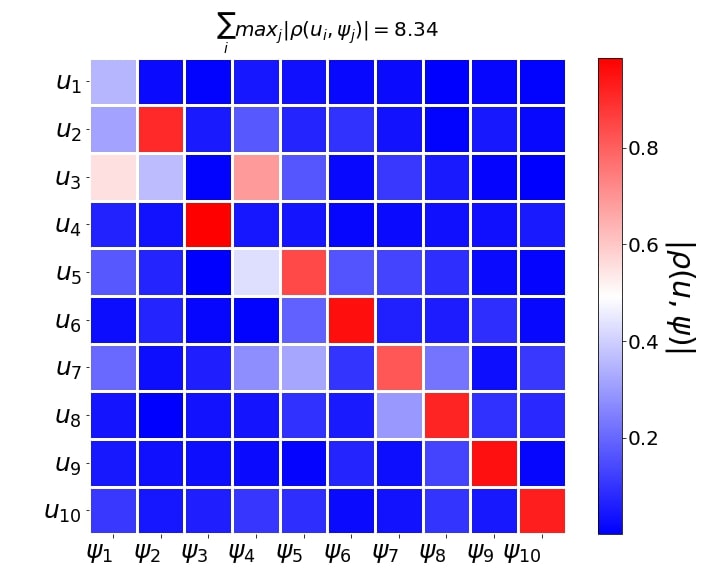}
	\caption{Word2vec embedding of samples from multiclass MNIST. We use $10000$ samples drawn from the MNIST handwritten dataset. The figure shows the absolute correlation between the $u_1,\ldots,u_{10}$ and $\psi_1,\ldots,\psi_{10}$. This results provide another empirical evidence that the eigenvectors of $P-\frac{1}{n}\bm 1$ are a good proxy for word2vec embedding. }
	\label{fig:highd2}
\end{figure}

\paragraph*{Acknowledgements}
The authors thank James Garritano and the anonymous reviewers for their helpful feedback. This work was supported by National Science Foundation DMS-1763179, Alfred P. Sloan Foundation.

\bibliographystyle{unsrt}
\bibliography{word2vec_bib}

\begin{thebibliography}{10}

\bibitem{mikolov2013efficient}
Tomas Mikolov, Kai Chen, Greg Corrado, and Jeffrey Dean.
\newblock Efficient estimation of word representations in vector space.
\newblock In Yoshua Bengio and Yann LeCun, editors, {\em 1st International
  Conference on Learning Representations, {ICLR} 2013, Scottsdale, Arizona,
  USA, May 2-4, 2013, Workshop Track Proceedings}, 2013.

\bibitem{goldberg2014word2vec}
Yoav Goldberg and Omer Levy.
\newblock word2vec explained: deriving mikolov et al.'s negative-sampling
  word-embedding method.
\newblock {\em arXiv preprint arXiv:1402.3722}, 2014.

\bibitem{grover2016node2vec}
Aditya Grover and Jure Leskovec.
\newblock node2vec: Scalable feature learning for networks.
\newblock In {\em Proceedings of the 22nd ACM SIGKDD international conference
  on Knowledge discovery and data mining}, pages 855--864, 2016.

\bibitem{mikolov2013distributed}
Tomas Mikolov, Ilya Sutskever, Kai Chen, Greg~S Corrado, and Jeff Dean.
\newblock Distributed representations of words and phrases and their
  compositionality.
\newblock In {\em Advances in neural information processing systems}, pages
  3111--3119, 2013.

\bibitem{le2014distributed}
Quoc Le and Tomas Mikolov.
\newblock Distributed representations of sentences and documents.
\newblock In {\em International conference on machine learning}, pages
  1188--1196, 2014.

\bibitem{narayanan2017graph2vec}
Annamalai Narayanan, Mahinthan Chandramohan, Rajasekar Venkatesan, Lihui Chen,
  Yang Liu, and Shantanu Jaiswal.
\newblock graph2vec: Learning distributed representations of graphs.
\newblock {\em arXiv preprint arXiv:1707.05005}, 2017.

\bibitem{hashimoto2016word}
Tatsunori~B Hashimoto, David Alvarez-Melis, and Tommi~S Jaakkola.
\newblock Word embeddings as metric recovery in semantic spaces.
\newblock {\em Transactions of the Association for Computational Linguistics},
  4:273--286, 2016.

\bibitem{hinton2003stochastic}
Geoffrey~E Hinton and Sam~T Roweis.
\newblock Stochastic neighbor embedding.
\newblock In {\em Advances in neural information processing systems}, pages
  857--864, 2003.

\bibitem{cotterell2017explaining}
Ryan Cotterell, Adam Poliak, Benjamin Van~Durme, and Jason Eisner.
\newblock Explaining and generalizing skip-gram through exponential family
  principal component analysis.
\newblock In {\em Proceedings of the 15th Conference of the European Chapter of
  the Association for Computational Linguistics: Volume 2, Short Papers}, pages
  175--181, 2017.

\bibitem{collins2002generalization}
Michael Collins, Sanjoy Dasgupta, and Robert~E Schapire.
\newblock A generalization of principal components analysis to the exponential
  family.
\newblock In {\em Advances in neural information processing systems}, pages
  617--624, 2002.

\bibitem{levy2014neural}
Omer Levy and Yoav Goldberg.
\newblock Neural word embedding as implicit matrix factorization.
\newblock In {\em Advances in neural information processing systems}, pages
  2177--2185, 2014.

\bibitem{qiu2018network}
Jiezhong Qiu, Yuxiao Dong, Hao Ma, Jian Li, Kuansan Wang, and Jie Tang.
\newblock Network embedding as matrix factorization: Unifying deepwalk, line,
  pte, and node2vec.
\newblock In {\em Proceedings of the Eleventh ACM International Conference on
  Web Search and Data Mining}, pages 459--467, 2018.

\bibitem{perozzi2014deepwalk}
Bryan Perozzi, Rami Al-Rfou, and Steven Skiena.
\newblock Deepwalk: Online learning of social representations.
\newblock In {\em Proceedings of the 20th ACM SIGKDD international conference
  on Knowledge discovery and data mining}, pages 701--710, 2014.

\bibitem{tang2015line}
Jian Tang, Meng Qu, Mingzhe Wang, Ming Zhang, Jun Yan, and Qiaozhu Mei.
\newblock Line: Large-scale information network embedding.
\newblock In {\em Proceedings of the 24th international conference on world
  wide web}, pages 1067--1077, 2015.

\bibitem{Arora2015RandomWO}
Sanjeev Arora, Yuanzhi Li, Yingyu Liang, Tengyu Ma, and Andrej Risteski.
\newblock Random walks on context spaces: Towards an explanation of the
  mysteries of semantic word embeddings.
\newblock {\em ArXiv}, abs/1502.03520, 2015.

\bibitem{landgraf2017word2vec}
Andrew~J Landgraf and Jeremy Bellay.
\newblock Word2vec skip-gram with negative sampling is a weighted logistic pca.
\newblock {\em arXiv preprint arXiv:1705.09755}, 2017.

\bibitem{belkin2003laplacian}
Mikhail Belkin and Partha Niyogi.
\newblock Laplacian eigenmaps for dimensionality reduction and data
  representation.
\newblock {\em Neural computation}, 15(6):1373--1396, 2003.

\bibitem{coifman2006diffusion}
Ronald~R Coifman and St{\'e}phane Lafon.
\newblock Diffusion maps.
\newblock {\em Applied and computational harmonic analysis}, 21(1):5--30, 2006.

\bibitem{singer2017spectral}
Amit Singer and Hau-Tieng Wu.
\newblock Spectral convergence of the connection laplacian from random samples.
\newblock {\em Information and Inference: A Journal of the IMA}, 6(1):58--123,
  2017.

\bibitem{belkin2007convergence}
Mikhail Belkin and Partha Niyogi.
\newblock Convergence of laplacian eigenmaps.
\newblock In {\em Advances in Neural Information Processing Systems}, pages
  129--136, 2007.

\bibitem{lafon2006data}
Stephane Lafon, Yosi Keller, and Ronald~R Coifman.
\newblock Data fusion and multicue data matching by diffusion maps.
\newblock {\em IEEE Transactions on pattern analysis and machine intelligence},
  28(11):1784--1797, 2006.

\bibitem{kernelscaling}
Ofir Lindenbaum, Moshe Salhov, Arie Yeredor, and Amir Averbuch.
\newblock Gaussian bandwidth selection for manifold learning and
  classification.
\newblock {\em Data Mining and Knowledge Discovery}, pages 1--37, 2020.

\bibitem{mnist}
Yann LeCun, L{\'e}on Bottou, Yoshua Bengio, and Patrick Haffner.
\newblock Gradient-based learning applied to document recognition.
\newblock {\em Proceedings of the IEEE}, 86(11):2278--2324, 1998.

\bibitem{coil100}
Sameer~A Nene, Shree~K Nayar, Hiroshi Murase, et~al.
\newblock Columbia object image library (coil-20).
\newblock 1996.

\bibitem{seismic}
Ofir Lindenbaum, Yuri Bregman, Neta Rabin, and Amir Averbuch.
\newblock Multiview kernels for low-dimensional modeling of seismic events.
\newblock {\em IEEE Transactions on Geoscience and Remote Sensing},
  56(6):3300--3310, 2018.

\bibitem{joswig}
Manfred Joswig.
\newblock Pattern recognition for earthquake detection.
\newblock {\em Bulletin of the Seismological Society of America},
  80(1):170--186, 1990.

\bibitem{alicetext}
Phillip Johnsen.
\newblock A text version of ``alice's adventures in wonderland''.
\newblock \url{https://gist.github.com/phillipj/4944029}, 2019 (accessed August
  8, 2020).

\end{thebibliography}

\appendix

\section{Relation between the examined functional and the Skip-gram model}\label{app:relation}

The skip-gram model was introduced in \cite{mikolov2013distributed} as a novel method for word embedding, given a text corpora. Let $x_1,\ldots,x_n$ be the vocabulary of the text, where every word $x_i$ appears in the text at least once. We denote by $p(x_j| x_i)$ the probability that  $x_j$ appears in a window of size $c$ around $x_i$. throughout the text. Let $y_k$ be the word in location $k$ in the text, where $k=1,\ldots,|\text{Text}|$.
The goal is to set the parameters $\theta\equiv \{w_i,v_i\}_{i=1}^n$ as to maximize a log probability function, defined for the pattern of neighborhoods in a window of size $c$,
\begin{equation}\label{eq:real_skipgram}
L(w,v) = \frac{1}{|\text{Text}|} \sum_{k=1}^{|\text{Text}|} \sum_{l \in [k-c,k+c]} \log (q(y_l| y_k; \theta)),
\end{equation}


where the conditional probability $q(x_i|x_j;\theta)$ is modeled using a soft-max function, 
\begin{eqnarray}\label{eq:model_probability_sg}
q(x_i|x_j; \theta) &=& \frac{\exp(\langle w_j, v_i\rangle )}{\sum_{m=1}^n \exp(\langle w_j, v_m\rangle ) }.
\end{eqnarray} 
By replacing the double summation in \eqref{eq:real_skipgram} with a double summation over the vocabulary yields
\begin{equation}\label{eq:sum_vocabulary}
L(w,v) = \sum_{i=1}^n\sum_{j=1}^n p(x_i)p(x_j|x_i)\log (q(x_j|x_i ; \theta))    ,
\end{equation}
where $p(x_i)$ is the prior on the word $x_i$.
Denoting $P_{ij}= p(x_i)p(x_j|x_i)$ and inserting  \eqref{eq:model_probability_sg} into \eqref{eq:sum_vocabulary} yields the function in \eqref{eq:skip_gram_function}. 
If we assume a uniform probability over the vocabulary, as in the experiments in sections \ref{sec:Examples} and \ref{sec:experiment high dim}, \eqref{eq:sum_vocabulary}
may be simplified,
\[
L(w,v) = \frac1n \sum_{i=1}^n\sum_{j=1}^n p(x_j|x_i)\log (q(x_j|x_i ; \theta)).
\]

\end{document}